\documentclass[lettersize,journal]{IEEEtran}
\usepackage{amsmath,amsfonts}
\usepackage{algorithmic}
\usepackage{algorithm}
\usepackage{array}
\usepackage[caption=false,font=normalsize,labelfont=sf,textfont=sf]{subfig}
\usepackage{textcomp}
\usepackage{stfloats}
\usepackage{url}
\usepackage{verbatim}
\usepackage{graphicx}
\usepackage{cite}
\usepackage{multirow}

\usepackage{makecell}
\usepackage{textcomp}
\usepackage{booktabs}
\usepackage{xcolor}
\newtheorem{theorem}{Theorem}

\def\BibTeX{{\rm B\kern-.05em{\sc i\kern-.025em b}\kern-.08em
    T\kern-.1667em\lower.7ex\hbox{E}\kern-.125emX}}
\begin{document}

\title{PRISM-AMC: Replay-Free Continual Learning for Automatic Modulation Classification}

\author{
    \IEEEauthorblockN{
        Usman Akram\IEEEauthorrefmark{1},
        Matthew Graham\IEEEauthorrefmark{2},
        and Haris Vikalo\IEEEauthorrefmark{1}
    }
    \\
    \IEEEauthorblockA{
        \IEEEauthorrefmark{1}Department of Electrical and Computer Engineering, University of Texas at Austin, Austin, TX, USA\\
        usman.akram@austin.utexas.edu, hvikalo@ece.utexas.edu
    }
    \\
    \IEEEauthorblockA{
        \IEEEauthorrefmark{2}Applied Research Laboratories, Austin, TX, USA\\
        matthew.graham@arlut.utexas.edu
    }
}
\maketitle

\begin{abstract}

Automatic modulation classification (AMC) is a key component of cognitive wireless systems, enabling spectrum awareness and adaptive communication. Most deep learning approaches to AMC assume a fixed set of modulation classes and access to previously observed training data. In practice, modulation classes and channel conditions may evolve over time, while storage or data-retention constraints may preclude replay of raw baseband I/Q samples. We propose PRISM-AMC, a replay-free continual learning framework that mitigates catastrophic forgetting by preserving task-relevant subspaces of intermediate representations via gradient projection memory. We further analyze projected stochastic-gradient updates under an SNR-dependent gradient-noise model. PRISM-AMC is evaluated in both class-incremental and domain-incremental settings. Experiments on the MIGOU and DeepSig 2018.01A datasets show substantially reduced forgetting relative to sequential fine-tuning and elastic weight consolidation, while remaining competitive with or outperforming replay-based RSD-AMC in class-incremental settings, without storing raw I/Q samples.

\end{abstract}


\section{Introduction}

Reliable identification of signal modulation plays a fundamental role in cognitive and software-defined radios operating in heterogeneous and uncoordinated spectrum environments, supporting the identification of unknown transmitters, dynamic spectrum access, and adaptive demodulation without prior signaling \cite{dobre2007survey,triantaris2019automatic}. Automatic modulation classification (AMC) supports blind demodulation and spectrum sharing in next-generation wireless networks \cite{ben2025automatic}. Recent advances in deep learning have significantly improved AMC performance by directly processing raw I/Q samples and learning robust representations under complex channel conditions \cite{tian2025survey}. Prior work has explored convolutional architectures, lightweight models, autoencoder-based methods, and federated learning for AMC \cite{zhang2019automatic,wang2020lightamc,9487492,wang2021federated,rahman2024improved,qi2022fedbkd}, illustrating the breadth of learning-based approaches.

\vspace{0.025in}

Despite recent progress, most deep learning approaches to AMC assume a fixed and fully known set of modulation classes available during training. In practice, modulation schemes may appear sequentially due to operational constraints or evolving deployment scenarios, requiring models to incorporate new classes or adapt to changing signal conditions while preserving performance on previously learned ones. This motivates a continual learning formulation of AMC, where the model encounters multiple tasks sequentially and must mitigate catastrophic forgetting.

\vspace{0.025in}

Early work on continual AMC has largely focused on class-incremental learning and relied on replay-based mechanisms \cite{yang2020modulation,zhao2023incremental}. However, replay requires retention or reconstruction of raw signal samples, which may be impractical under storage, data-governance, or regulatory constraints. In many deployments, retention or sharing of baseband I/Q data may be restricted, necessitating replay-free adaptation mechanisms.

\vspace{0.025in}

To address this challenge, we propose PRISM-AMC, a replay-free continual learning framework based on gradient projection memory (GPM) that mitigates forgetting without storing raw I/Q samples. Instead, PRISM-AMC retains low-dimensional subspace representations derived from intermediate activations and constrains subsequent updates to avoid interference with prior tasks. Experiments on simulated and over-the-air datasets demonstrate reduced forgetting relative to sequential fine-tuning and EWC, while achieving competitive or superior performance relative to replay-based baselines in the class-incremental settings.

\vspace{0.025in}

The contributions of this work can be summarized as follows: (i) we formulate automatic modulation classification as a continual learning problem under raw-sample retention constraints; (ii) we develop a replay-free framework based on gradient projection memory that mitigates catastrophic forgetting without storing raw I/Q samples; (iii) we analyze projected Randomized Stochastic Gradient (RSG) updates \cite{ghadimi2013stochasticfirstzerothordermethods} under an SNR-dependent gradient-noise model and derive a bound on the expected squared norm of the projected gradient; (iv) we provide empirical evaluations on over-the-air and synthetic datasets demonstrating reduced forgetting and competitive or improved accuracy relative to sequential, regularization-based, and replay-based baselines; and (v) we also investigate a domain-incremental setting where the tasks differ in the underlying channel conditions rather than the modulation classes.\footnote{A preliminary version of this work appeared at FLICS 2026 \cite{11621924}. This journal manuscript substantially extends the conference paper with new theoretical analysis and expanded class- and domain-incremental evaluations.}

\label{sec:intro}



\section{System Model}
\label{sysmod}

We consider the problem of identifying the modulation scheme underlying a received complex baseband signal. The received samples over an observation window of length $N$ are modeled as
\begin{align}
r[n] = A e^{j\left(\frac{2\pi \Delta f n}{N} + \Delta \theta\right)} s[n] + w[n], \label{sysmodeq}
\end{align}
where $s[n]$ denotes a transmitted sample drawn according to modulation scheme $\Omega_c \in \{\Omega_1,\ldots,\Omega_C\}$, $A$ and $\Delta\theta$ denote amplitude and phase offsets, $\Delta f$ is the carrier frequency offset (CFO), and $w[n]$ is complex Gaussian noise. The task is to infer $\Omega_c$ from the received I/Q samples over the observation window.

\vspace{0.025in}

In this work, we study continual learning for automatic modulation classification in a sequence of tasks whose modulation classes or underlying signal distributions may change over time. At task $t$, the model is trained on a dataset $\mathcal{D}_t$ associated with a set of modulation schemes $\Omega^{t}=\{\Omega^{t}_1,\ldots,\Omega^{t}_{C_t}\}$, where $C_t$ is the number of modulation classes present in task $t$. The model is sequentially trained on datasets $\mathcal{D}_1,\ldots,\mathcal{D}_T$, each corresponding to modulation sets $\Omega^{1},\ldots,\Omega^{T}$, respectively. In the class-incremental setting, new modulation classes are introduced over time, and the objective is to learn them while preserving performance on previously observed classes. In the domain-incremental setting, the modulation classes remain fixed across tasks, while the underlying channel conditions and signal statistics change.

\section{PRISM-AMC Framework}
\label{propmeth}

To mitigate catastrophic forgetting under raw-sample retention constraints, we develop PRISM-AMC, a projection-based continual learning framework for sequentially encountered raw I/Q data. Unlike replay-based approaches, PRISM-AMC neither stores nor regenerates previously observed baseband samples. Instead, it preserves task-relevant subspaces of intermediate feature representations and projects subsequent gradient updates onto their orthogonal complement, thereby reducing interference with previously learned tasks.

\vspace{0.025in}

PRISM-AMC builds upon the Gradient Projection Memory (GPM) framework \cite{sahagradient}, which exploits the fact that the gradient with respect to a neural network layer's weights lies in the span of its input activations. By identifying task-specific activation subspaces and projecting future gradients onto the orthogonal complement of retained subspaces from prior tasks, GPM reduces interference between sequential tasks. In PRISM-AMC, this principle is applied to intermediate feature representations of the modulation classifier to preserve prior knowledge without storing raw signal samples. For convolutional layers, the same property holds after unfolding the convolution into matrix form. Let $X$ denote the matrix of input feature patches and $\Delta$ the backpropagated error. The gradient with respect to the filter weights can be expressed as
\[
\nabla_W L = X^{\top} \Delta,
\]
so its columns lie in the subspace spanned by the input feature patches. Thus, activation subspaces can be estimated for convolutional layers and used to constrain future gradients in the same manner as for fully connected layers.

\vspace{0.025in}

The central idea of gradient projection memory is to decompose the gradient space at each layer into two orthogonal subspaces: the Core Gradient Space (CGS) and the Residual Gradient Space (RGS). After training on task $1$ with dataset $\mathcal{D}_1$, we select $n_s$ samples from $\mathcal{D}_1$ and construct, for each layer $l$, a representation matrix $\mathbf{R}_1^l$ using the corresponding input activations. For a fully connected layer,
\[
\mathbf{R}_1^l = [\mathbf{x}_{1,1}^l,\ldots,\mathbf{x}_{n_s,1}^l],
\]
and for a convolutional layer, the unfolded feature patches are stacked similarly. To obtain a basis for the CGS, we perform singular value decomposition (SVD),
$\mathbf{R}_1^l = U_1^l \Sigma_1^l (V_1^l)^{\top}$,
and select the smallest rank $k_1^l$ such that
\[
\|(\mathbf{R}_1^l)_{k_1^l}\|_F^2
\ge
\epsilon_{\mathrm{th}}^l \|\mathbf{R}_1^l\|_F^2,
\]
where $(\mathbf{R}_1^l)_{k_1^l}$ denotes the rank-$k_1^l$ truncated SVD approximation. The CGS for layer $l$, denoted $S^l$, is initialized as the span of the first $k_1^l$ columns of $U_1^l$. The corresponding basis vectors are stored in the gradient projection memory $\mathcal{M}=(M^l)_{l=1}^L$, where
\[
M^l=[u_{1,1}^l,\ldots,u_{k_1^l,1}^l].
\]

\vspace{0.025in}

For tasks $t>1$, the gradients are projected onto the RGS to reduce interference with previously learned tasks:
\begin{align}
\text{FC layer:}\quad
\widetilde{\nabla}_{W_t^l} L_t
&=
\nabla_{W_t^l} L_t
- \nabla_{W_t^l} L_t M^l (M^l)^{\top},
\label{eq_proj_1}\\
\text{Conv layer:}\quad
\widetilde{\nabla}_{W_t^l} L_t
&=
\nabla_{W_t^l} L_t
- M^l (M^l)^{\top} \nabla_{W_t^l} L_t.
\label{eq_proj_2}
\end{align}

At the end of task $t$, activations are collected to form $\mathbf{R}_t^l$, which is projected onto the RGS:
\begin{align}
\hat{\mathbf{R}}_t^l
=
\mathbf{R}_t^l - M^l (M^l)^{\top} \mathbf{R}_t^l.
\label{eq_updatemem_1}
\end{align}
Performing SVD on
$\hat{\mathbf{R}}_t^l
=
\hat{U}_t^l \hat{\Sigma}_t^l (\hat{V}_t^l)^{\top}$,
we select the smallest $k_t^l$ such that
\[
\|\mathbf{R}^{l}_{t,\mathrm{proj}}\|_F^2
+
\|(\hat{\mathbf{R}}_t^l)_{k_t^l}\|_F^2
\ge
\epsilon_{\mathrm{th}}^l \|\mathbf{R}_t^l\|_F^2,
\]
where
\[
\mathbf{R}^{l}_{t,\mathrm{proj}}
=
M^l (M^l)^{\top} \mathbf{R}_t^l
\]
is the component of the current-task representation already captured by the retained CGS. The basis $M^l$ is then updated by appending the first $k_t^l$ columns of $\hat{U}_t^l$. The complete training procedure is summarized in Algorithm~\ref{alg:gpm}.

\begin{algorithm}[t]
\caption{\textbf{PRISM-AMC Training}}
\label{alg:gpm}
\begin{algorithmic}[1]
\STATE \textbf{Input:} Model $f_W$, task datasets $\{\mathcal{D}_t\}_{t=1}^T$, learning rate $\eta$, thresholds $\{\epsilon_{\mathrm{th}}^l\}_{l=1}^L$, loss $\mathcal{L}$
\STATE Initialize $M^{l} \gets [\ ]$ for $l=1,\ldots,L$
\STATE $\mathcal{M} \gets (M^l)_{l=1}^L$, $W \gets W_{0}$

\FOR{$t = 1,\ldots,T$}
    \REPEAT
        \STATE Sample mini-batch $B \sim \mathcal{D}_t$
        \STATE $\Delta_W \gets \operatorname{Gradients}(B, f_W, \mathcal{L})$
        \IF{$t > 1$}
            \STATE $\Delta_W \gets \operatorname{PROJ}(\Delta_W, \mathcal{M})$
        \ENDIF
        \STATE $W \gets W - \eta \Delta_W$
    \UNTIL{convergence}

    \STATE Sample $n_s$ examples $B_{n_s} \sim \mathcal{D}_t$
    \STATE Compute layer-wise representation matrices $\{\mathbf{R}_t^l\}_{l=1}^L$

    \FOR{$l = 1,\ldots,L$}
        \IF{$M^l$ not empty}
            \STATE $\mathbf{R}_{t,\mathrm{proj}}^l
            \gets M^l (M^l)^{\top} \mathbf{R}_t^l$, 
            $\hat{\mathbf{R}}_t^l
            \gets \mathbf{R}_t^l-\mathbf{R}_{t,\mathrm{proj}}^l$
        \ELSE
            \STATE $\mathbf{R}_{t,\mathrm{proj}}^l \gets 0$, 
            $\hat{\mathbf{R}}_t^l \gets \mathbf{R}_t^l$
        \ENDIF

        \STATE $\hat{U}_t^l \gets \operatorname{SVD}(\hat{\mathbf{R}}_t^l)$
        \STATE Select the smallest $k_t^l$ such that $\|\mathbf{R}_{t,\mathrm{proj}}^l\|_F^2
        +
        \|(\hat{\mathbf{R}}_t^l)_{k_t^l}\|_F^2
        \ge
        \epsilon_{\mathrm{th}}^l \|\mathbf{R}_t^l\|_F^2
        $
        \STATE $M^l \gets
        [\,M^l,\ \hat{U}_t^l(:,1\!:\!k_t^l)\,]$
    \ENDFOR
\ENDFOR

\STATE \textbf{return} $f_W$
\color{white}
\end{algorithmic}
\end{algorithm}

\vspace{0.025in}

The original GPM framework was developed for convolutional and fully connected layers \cite{sahagradient}. Accordingly, PRISM-AMC uses a lightweight 1D convolutional neural network for modulation classification. Such architectures have demonstrated strong performance on raw I/Q data in prior AMC literature \cite{wang2020lightamc,wang2021federated}. The network architecture is summarized in Table~\ref{tab:cnn1d_arch}. We omit bias terms in the lower layers, which we found to improve performance in the continual learning setting.

\section{Analysis of Projected Stochastic Updates}

In this section, we analyze the optimization behavior induced by the
gradient-projection mechanism used in PRISM-AMC. Specifically, we consider
the stochastic-gradient updates within a task while holding the retained
subspace from previous tasks fixed. Relying on the Randomized Stochastic Gradient (RSG) framework \cite{ghadimi2013stochasticfirstzerothordermethods}, we derive
a bound on the expected squared norm of the gradient projected onto the
residual subspace under an SNR-dependent gradient-noise model. Let $f:\mathbb{R}^n\rightarrow\mathbb{R}$ be a
differentiable, not necessarily convex function that is bounded below and has
an $L$-Lipschitz continuous gradient, i.e.,
\begin{align}
\|\nabla f(y)-\nabla f(x)\|
\leq L\|y-x\|,
\qquad \forall x,y\in\mathbb{R}^n.
\end{align}
It then follows that
\begin{align}
\left|
f(y)-f(x)-\langle \nabla f(x),y-x\rangle
\right|
\leq \frac{L}{2}\|y-x\|^2,
\; \forall x,y\in\mathbb{R}^n.
\end{align}
Consider the unconstrained stochastic optimization problem
\[
f^*=\inf_{x\in\mathbb{R}^n}f(x).
\]
At iteration $k$, a stochastic first-order oracle returns a gradient estimate
$G(x_k,\xi_k)$ satisfying
\begin{align}
\mathbb{E}\!\left[G(x_k,\xi_k)\mid x_k\right]
&= \nabla f(x_k),\\
\mathbb{E}\!\left[
\|G(x_k,\xi_k)-\nabla f(x_k)\|^2
\mid x_k
\right]
&\leq \sigma^2+\alpha\gamma^{-1},
\label{grad_ineq}
\end{align}
where $\gamma$ denotes the (linear scale) signal-to-noise ratio (SNR),
$\alpha\geq 0$ controls the SNR-dependent contribution to the gradient
variance, and $\sigma^2$ accounts for SNR-independent stochasticity arising
from factors such as mini-batch sampling. In our prior work
\cite{akram2025federatedselfsupervisedlearningautomatic}, we derived an
additive SNR-dependent term in a gradient-variance bound for self-supervised
learning of wireless signals, motivating the variance model in
\eqref{grad_ineq}. 

To formalize the within-task PRISM-AMC update, we vectorize the network
parameters and their gradients. Each layerwise projection in
\eqref{eq_proj_1}--\eqref{eq_proj_2} induces an orthogonal projection on the
corresponding vectorized parameter block. Collecting these projections across
layers yields a block-diagonal orthogonal projector $\mathcal{P}$ on the full
parameter space. Equivalently, we write
\[
\mathcal{P}=I-MM^\top,
\]
where the columns of $M$ form an orthonormal basis for the retained directions
associated with prior tasks in the vectorized parameter space. During
optimization of a given task, these retained directions are fixed, and
$\mathcal{P}$ projects the stochastic gradient onto the corresponding residual
subspace. For the analysis, we consider the projected RSG iteration in
Algorithm~\ref{alg:PRSG}.

\begin{algorithm}[t]
\caption{Projected Randomized Stochastic Gradient (RSG) Method}
\label{alg:PRSG}
\begin{algorithmic}[1]
\REQUIRE Initial point $x_1$, iteration limit $N$, stepsizes
$\{\eta_k\}_{k\geq 1}$, probability mass function $P_R(\cdot)$
supported on $\{1,\dots,N\}$, orthogonal projection matrix $\mathcal{P}$
\STATE Generate a random variable $R$ with probability mass function $P_R$
\FOR{$k = 1,\ldots,R$}
    \STATE Call the stochastic first-order oracle to obtain $G(x_k,\xi_k)$
    \STATE Update
    \[
    x_{k+1}
    =
    x_k-\eta_k\mathcal{P}G(x_k,\xi_k)
    \]
\ENDFOR
\STATE \textbf{return} $x_R$
\color{white}
\end{algorithmic}
\end{algorithm}

The following theorem places an upper bound on the expected squared norm of
the projected gradient.

\begin{theorem}
Suppose that $\mathcal{P}$ is a fixed orthogonal projection matrix and that
the stepsizes $\{\eta_k\}$ and the probability mass function $P_R(\cdot)$
in the RSG method are chosen such that
$0<\eta_k<\frac{2}{L}$,
and
\begin{equation}
P_R(k)
:=
\mathrm{Prob}\{R=k\}
=
\frac{2\eta_k-L\eta_k^2}
{\sum_{j=1}^N(2\eta_j-L\eta_j^2)},
\; k=1,\dots,N.
\end{equation}
Assume that $R$ is independent of the stochastic oracle randomness
$\xi_{[N]}=(\xi_1,\ldots,\xi_N)$. Then, for any $N\geq1$,
\begin{equation}
\frac{1}{L}
\mathbb{E}\!\left[\|\mathcal{P}\nabla f(x_R)\|^2\right]
\leq
\frac{
D_f^2+
(\sigma^2+\alpha\gamma^{-1})
\sum_{k=1}^N\eta_k^2
}{
\sum_{k=1}^N(2\eta_k-L\eta_k^2)
},
\end{equation}
where the expectation is taken with respect to $R$ and the stochastic oracle
randomness, and
\begin{equation}
D_f
:=
\left[
\frac{2(f(x_1)-f^*)}{L}
\right]^{1/2}.
\end{equation}
\end{theorem}

\noindent\textbf{Proof:}
The proof follows the RSG analysis in
\cite{ghadimi2013stochasticfirstzerothordermethods}, modified to account for
the projected stochastic-gradient update. Let
\[
G_k := G(x_k,\xi_k),
\qquad
g_k := \nabla f(x_k).
\]
Since $\mathcal{P}$ is an orthogonal projection matrix,
$\mathcal{P}^{\top}=\mathcal{P}$ and $\mathcal{P}^2=\mathcal{P}$. Using the $L$-Lipschitz continuity of $\nabla f$ and the update
\[
x_{k+1}=x_k-\eta_k\mathcal{P}G_k,
\]
we have
\begin{align}
f(x_{k+1})
&\leq
f(x_k)
-\eta_k \langle g_k,\mathcal{P}G_k\rangle
+\frac{L}{2}\eta_k^2\|\mathcal{P}G_k\|^2.
\label{eq:proj_descent}
\end{align}

Let $\mathbb{E}_k[\cdot]$ denote expectation with respect to $\xi_k$
conditioned on $x_k$. By the unbiasedness of the stochastic first-order
oracle,
\begin{align}
\mathbb{E}_k[\langle g_k,\mathcal{P}G_k\rangle]
=
\langle g_k,\mathcal{P}g_k\rangle =
\|\mathcal{P}g_k\|^2.
\end{align}
Moreover,
\begin{align}
\mathbb{E}_k[\|\mathcal{P}G_k\|^2]
&=
\|\mathcal{P}g_k\|^2
+
\mathbb{E}_k\!\left[
\|\mathcal{P}(G_k-g_k)\|^2
\right] \nonumber\\
&\leq
\|\mathcal{P}g_k\|^2
+
\sigma^2+\alpha\gamma^{-1},
\end{align}
where the inequality follows from
$\|\mathcal{P}v\|\leq\|v\|$ for an orthogonal projector and
\eqref{grad_ineq}.

Taking conditional expectation in \eqref{eq:proj_descent} therefore gives
\begin{eqnarray}
\mathbb{E}_k[f(x_{k+1})]
&\leq&
f(x_k)
-
\left(
\eta_k-\frac{L}{2}\eta_k^2
\right)
\|\mathcal{P}g_k\|^2 \nonumber \\
&+&
\frac{L}{2}\eta_k^2
(\sigma^2+\alpha\gamma^{-1}).
\end{eqnarray}
Taking total expectation, summing over $k=1,\ldots,N$, and using
$f(x_{N+1})\geq f^*$ yields
\begin{align}
\sum_{k=1}^N
\left(
\eta_k-\frac{L}{2}\eta_k^2
\right)
\mathbb{E}\!\left[
\|\mathcal{P}\nabla f(x_k)\|^2
\right]
\leq \nonumber \\
f(x_1)-f^*
+
\frac{L}{2}
(\sigma^2+\alpha\gamma^{-1})
\sum_{k=1}^N \eta_k^2.
\label{eq:proj_sum_bound}
\end{align}

By the definition of $P_R(k)$ and the independence of $R$ from the stochastic
oracle randomness,
\begin{align}
\mathbb{E}\!\left[
\|\mathcal{P}\nabla f(x_R)\|^2
\right]
=
\frac{
\sum_{k=1}^N
(2\eta_k-L\eta_k^2)
\mathbb{E}\!\left[
\|\mathcal{P}\nabla f(x_k)\|^2
\right]
}{
\sum_{k=1}^N
(2\eta_k-L\eta_k^2)
}.
\end{align}
Multiplying \eqref{eq:proj_sum_bound} by $2$ and substituting the above
identity gives
\begin{align}
\frac{1}{L}
\mathbb{E}\!\left[
\|\mathcal{P}\nabla f(x_R)\|^2
\right]
&\leq
\frac{
\frac{2(f(x_1)-f^*)}{L}
+
(\sigma^2+\alpha\gamma^{-1})
\sum_{k=1}^N\eta_k^2
}{
\sum_{k=1}^N
(2\eta_k-L\eta_k^2)
}.
\end{align}
Using
\[
D_f^2=\frac{2(f(x_1)-f^*)}{L}
\]
establishes the stated result.
\hfill$\square$

With an appropriate choice of stepsizes, Theorem~1 yields
\[
\mathbb{E}\left[\|\mathcal{P}\nabla f(x_R)\|^2\right]
=
\mathcal{O}\left(\frac{1}{N}\right)
+
\mathcal{O}\left(
\sqrt{\frac{\sigma^2+\alpha\gamma^{-1}}{N}}
\right).
\]
Thus, for any fixed SNR $\gamma>0$,
$\mathbb{E}\left[\|\mathcal{P}\nabla f(x_R)\|^2\right]
\rightarrow 0$ as $N\rightarrow\infty$.
This rate is formalized in the following theorem.

\begin{theorem}
Under the assumptions of Theorem~1, suppose that the stepsizes
$\{\eta_k\}$ are set to
\begin{equation}
\eta_k
=
\min \left(
\frac{1}{L},
\frac{\tilde{D}}
{\sqrt{(\sigma^2+\alpha\gamma^{-1})N}}
\right),
\quad k=1,\dots,N,
\end{equation}
for some $\tilde{D}>0$, with
$\sigma^2+\alpha\gamma^{-1}>0$. Let the probability mass function
$P_R(\cdot)$ be given by
\begin{equation}
P_R(k)
=
\frac{2\eta_k-L\eta_k^2}
{\sum_{j=1}^N(2\eta_j-L\eta_j^2)},
\quad k=1,\dots,N.
\end{equation}
Then,
\begin{equation}
\frac{1}{L}
\mathbb{E}\left[\|\mathcal{P}\nabla f(x_R)\|^2\right]
\leq
\frac{L D_f^2}{N}
+
\left(
\tilde{D}+\frac{D_f^2}{\tilde{D}}
\right)
\sqrt{\frac{\sigma^2+\alpha\gamma^{-1}}{N}},
\end{equation}
where
\begin{equation}
D_f
:=
\left[
\frac{2(f(x_1)-f^*)}{L}
\right]^{1/2}.
\end{equation}
\end{theorem}

\noindent\textbf{Proof:}
Let
\[
v:=\sigma^2+\alpha\gamma^{-1},
\qquad
\eta
:=
\min\left\{
\frac{1}{L},
\frac{\tilde{D}}{\sqrt{vN}}
\right\}.
\]
Since $\eta_k=\eta$ for all $k$, Theorem~1 gives
\begin{align}
\frac{D_f^2+v\sum_{k=1}^N\eta_k^2}
{\sum_{k=1}^N(2\eta_k-L\eta_k^2)}
&=
\frac{D_f^2+Nv\eta^2}
{N\eta(2-L\eta)}.
\end{align}
Because $\eta\leq 1/L$, we have $2-L\eta\geq 1$, and hence
\begin{align}
\frac{D_f^2+Nv\eta^2}
{N\eta(2-L\eta)}
&\leq
\frac{D_f^2}{N\eta}+v\eta.
\end{align}
From the definition of $\eta$,
\[
\frac{1}{\eta}
=
\max\left\{
L,
\frac{\sqrt{vN}}{\tilde{D}}
\right\},
\]
and
\[
\eta
\leq
\frac{\tilde{D}}{\sqrt{vN}}.
\]
Therefore,
\begin{align}
\frac{D_f^2}{N\eta}+v\eta
&\leq
\frac{D_f^2}{N}
\max\left\{
L,
\frac{\sqrt{vN}}{\tilde{D}}
\right\}
+
\tilde{D}\sqrt{\frac{v}{N}}
\nonumber\\
&\leq
\frac{LD_f^2}{N}
+
\left(
\tilde{D}+\frac{D_f^2}{\tilde{D}}
\right)
\sqrt{\frac{v}{N}}.
\end{align}
Substituting $v=\sigma^2+\alpha\gamma^{-1}$ establishes the result.
\hfill$\square$

\section{Experimental Results}
\label{Exp_Res}

For our experiments, we evaluate PRISM-AMC against the following baselines:
\begin{itemize}
\item \textbf{Sequential Fine-Tuning (SFT):} The model is trained sequentially across tasks without any mechanism to mitigate catastrophic forgetting.

\item \textbf{Replay and Soft Distillation (RSD-AMC):} An adaptation of the replay-based class-incremental framework proposed in \cite{yang2020modulation}, which combines replay of previously observed samples with knowledge distillation.

\item \textbf{Elastic Weight Consolidation (EWC):}
A replay-free continual learning baseline \cite{kirkpatrick2017overcoming} that mitigates catastrophic forgetting by penalizing deviations from previously learned parameters. After completing task $k$, we store the parameters $\theta^{*(k)}$ and estimate their importance via the diagonal empirical Fisher information matrix $F^{(k)}$. During training on subsequent tasks, we add a quadratic regularization term
\begin{equation}
\mathcal{L}_{\text{EWC}}
= \frac{\lambda_{\text{EWC}}}{2} \sum_{k < t} \sum_i F_i^{(k)}
\left( \theta_i - \theta_i^{*(k)} \right)^2.
\end{equation}

\item \textbf{MSNCIL:}
A task-incremental method proposed in \cite{10980115} that avoids storing
historical training data by partitioning the network into task-specific
submodels via neural state assignment. Once assigned to a task, the
corresponding neurons are frozen during subsequent training, yielding
a BWT of $0$ by construction in the task-incremental setting. We implement
MSNCIL using the hyperparameter configuration reported in
\cite{10980115}. Since MSNCIL assumes task-specific subnetworks and
task identity at inference time, we do not include it in the strict
single-head class-incremental experiments.

\item \textbf{PASS-Net:}
For the strict single-head class-incremental setting, we consider PASS-Net,
a few-shot class-incremental learning method proposed in \cite{Passnet} that
leverages pseudo-classes and stochastic classifiers for AMC. We implement
PASS-Net using the architecture and relevant hyperparameters reported in
\cite{Passnet}.

\end{itemize}

Unless specified otherwise, the baselines employ the same network architecture
as PRISM-AMC for a fair comparison; PASS-Net uses the architecture reported
in \cite{Passnet}. We report final average accuracy (ACC) and backward transfer
(BWT). The final average accuracy is defined as
\[
\mathrm{ACC}
=
\frac{1}{T}\sum_{t=1}^{T} acc_{T,t},
\]
and BWT is defined as
\[
\mathrm{BWT}
=
\frac{1}{T-1}
\sum_{t=1}^{T-1}
\left(acc_{T,t}-acc_{t,t}\right),
\]
where $acc_{j,t}$ denotes the accuracy on task $t$ after training up to task
$j$. More negative BWT values indicate greater forgetting. Unless specified
otherwise, all methods use separate task-specific classifier heads.

\vspace{0.025in}

Unless otherwise specified, PRISM-AMC, SFT, RSD-AMC, and EWC are trained
for 70 epochs per task using Adam with an initial learning rate
$\eta_{\text{init}} = 0.001$. A ReduceLROnPlateau scheduler reduces the
learning rate by a factor of 0.5 with patience of 6 epochs. For PRISM-AMC,
the energy threshold $\epsilon_{\mathrm{th}}$ is initialized to 0.97 for all
layers and increased by 0.003 after each task, and the activation batch size
for subspace estimation is fixed to $n_s = 125$. For RSD-AMC, the replay
memory is limited to 125 samples per task with approximately uniform class
representation. For EWC, we set $\lambda_{\mathrm{EWC}}$ to 500.

\begin{table}[!htbp]
\centering
\caption{1D CNN architecture (input length $N$)}
\label{tab:cnn1d_arch}
\resizebox{\linewidth}{!}{%
\begin{tabular}{c c c c c}
\toprule
\textbf{Layer} & \textbf{Type} & \textbf{Kernel / Units} & \textbf{Output} & \textbf{Details} \\
\midrule
Input & -- & -- & $2 \times N$ & I/Q samples \\

Conv1 & 1D Conv & $128 \times 16$ & $128 \times (N+1)$ &
stride 1, pad 8, no bias \\

BN1 + ReLU & BN + ReLU & -- & $128 \times (N+1)$ &
Dropout 0.1 \\

Conv2 & 1D Conv & $64 \times 8$ & $64 \times (N+2)$ &
stride 1, pad 4, no bias \\

BN2 + ReLU & BN + ReLU & -- & $64 \times (N+2)$ &
Dropout 0.1 \\

Flatten & -- & -- & $64(N+2)$ & -- \\

FC1 & FC & 256 & 256 & no bias \\

BN3 + ReLU & BN + ReLU & -- & 256 & Dropout 0.5 \\

FC2 & FC & 128 & 128 & no bias \\

BN4 + ReLU & BN + ReLU & -- & 128 & Dropout 0.5 \\

FC3$_t$ & Task Head & $n_t$ & $n_t$ &
task-specific-head settings \\

\bottomrule
\end{tabular}
}
\end{table}

\subsection{Results on MIGOU}
\subsubsection{Task-specific-head settings}

We first evaluate PRISM-AMC on the MIGOU dataset~\cite{utrilla2019migou},
which contains over-the-air I/Q measurements from 11 modulation classes
(e.g., AM, FM, QAM, PSK) transmitted using a USRP B210 and recorded at
distances of 1\,m and 6\,m, corresponding to average SNRs of 37\,dB and
22\,dB. Each sample consists of $N=128$ I/Q symbols. We consider $T=5$
tasks, each containing 3 randomly selected modulation classes with
independent label assignments, allowing identical modulation types to appear
in different tasks with task-specific labels. For each task, 35,000 samples
per class are used for training, and 7,000 per class for each of validation and testing.

Tables~\ref{tab:cl_matrix_prism}--\ref{tab:migou_cl_comparison} report the continual learning performance for PRISM-AMC and baseline algorithms. PRISM-AMC achieves performance competitive with RSD-AMC and MSNCIL while outperforming EWC and SFT in terms of final average accuracy by 4.88 and 60.00 percentage points, respectively. The improvement is more pronounced on early tasks evaluated after training on all tasks, reflecting stronger retention. This is further corroborated by BWT improvements of 6.15 and 75.03 percentage points over EWC and SFT, respectively.

We additionally consider a strict class-disjoint setting in which Task~1
contains three modulation classes and each subsequent task introduces two
previously unseen classes. As shown in Table~\ref{tab:migounooverlap},
PRISM-AMC maintains competitive average accuracy and achieves the best BWT
among the competing baselines other than MSNCIL. MSNCIL achieves a BWT of
zero by construction, but has lower average accuracy than both PRISM-AMC and
RSD-AMC. A drop in accuracy on Task~5 exhibited by all methods may reflect
the difficulty of distinguishing AM-SSB and AM-DSB in the I/Q domain. This
is consistent with the fact that even sequential fine-tuning (SFT), which
imposes no constraints on model updates, exhibits reduced performance on
Task~5. For PRISM-AMC, the reduction in Task~5 accuracy may also be influenced
by the accumulation of subspace constraints across tasks. As more tasks are
learned, the residual gradient space available for adaptation shrinks, which
can limit the model's ability to learn fine-grained distinctions in later
tasks. As demonstrated by the ablation experiments on
$\epsilon_{\mathrm{th}}$ presented later, this reflects the inherent
stability--plasticity tradeoff: preserving performance on earlier tasks
reduces flexibility for adapting to more subtle and challenging new tasks.

Training time per epoch is summarized in Table~\ref{tab:time_per_epoch}.
Replay-based training grows from 68\,s (Task~1) to 216\,s (Task~5) due to
the expanding replay buffer, while SFT remains nearly constant at
approximately 34\,s after Task~1. The compute time for Elastic Weight
Consolidation (EWC) increases with the number of tasks since the quadratic
regularization term must be computed with respect to task-specific Fisher
information of all previously learned tasks. MSNCIL remains approximately
constant at 42\,s per epoch from Tasks~2 through~5. PRISM-AMC incurs
additional per-epoch cost from gradient projection beginning with Task~2,
but stabilizes at roughly 86\,s per epoch thereafter. In these experiments,
its per-epoch training time remains essentially constant from Tasks~2
through~5.

Memory consumption across tasks is reported in
Table~\ref{tab:memory_growth_migou}. PRISM-AMC stores only the retained
subspace bases derived from intermediate activations, rather than raw I/Q
samples, and requires memory comparable to the size of the model. In the
considered small-replay regime, RSD-AMC uses less memory, but does so by
retaining raw I/Q samples; its storage grows with the replay-buffer size,
the number of tasks, and the signal dimensionality. MSNCIL maintains a
constant memory footprint of approximately 8.52\,MB across tasks. EWC has
the largest memory footprint because it stores task-specific parameter
snapshots together with the corresponding diagonal Fisher information.
For each task, these quantities together require approximately twice the
model size, so the total EWC memory grows approximately linearly with the
number of tasks.

Next, we conduct an ablation study on $\epsilon_{\mathrm{th}}$, which controls the amount of representation energy retained in the stored subspace and hence the stability--plasticity tradeoff. Smaller $\epsilon_{\mathrm{th}}$ implies fewer basis vectors representing prior tasks, thus imposing less restriction on learning a new task; this, however, increases the chance of forgetting. Table~\ref{tab:gpm_eps} further illustrates this tradeoff; as can be seen in the table, accuracy on the earliest task, namely Task~1, generally increases with $\epsilon_{\mathrm{th}}$, whereas accuracy on the latest task, Task~5, generally decreases. Consequently, BWT generally improves as $\epsilon_{\mathrm{th}}$ grows. Finally, the ablation results presented in Table~\ref{tab:bias} support the omission of bias terms in the lower layers.

\begin{table}[!htbp]
\centering
\footnotesize
\caption{Continual learning performance on the MIGOU dataset using PRISM-AMC:
task--task accuracy matrix. Entry $(i,j)$ denotes the accuracy on task $j$
after training up to task $i$. Final average accuracy: $99.20\%$; BWT:
$-0.87$ percentage points.}
\label{tab:cl_matrix_prism}
\renewcommand{\arraystretch}{1.1}
\setlength{\tabcolsep}{4pt}
\begin{tabular}{|c|c|c|c|c|c|}
\hline
 & \multicolumn{5}{c|}{\textbf{Evaluation Task}} \\
\hline
\textbf{After Task} & \textbf{T1} & \textbf{T2} & \textbf{T3} & \textbf{T4} & \textbf{T5} \\
\hline
\textbf{T1} & 99.9 & -- & -- & -- & -- \\
\hline
\textbf{T2} & 92.6 & 99.9 & -- & -- & -- \\
\hline
\textbf{T3} & 95.3 & 99.4 & 100.0 & -- & -- \\
\hline
\textbf{T4} & 99.0 & 97.4 & 93.3 & 99.7 & -- \\
\hline
\textbf{T5} & 99.1 & 99.0 & 99.9 & 98.0 & 100.0 \\
\hline
\end{tabular}
\end{table}

\begin{table}[!t]
\centering
\caption{Continual learning performance on the MIGOU dataset across five tasks with three classes per task.
Reporting final task accuracies after training on all five tasks,
overall average accuracy (Avg), and backward transfer (BWT).}
\label{tab:migou_cl_comparison}
\resizebox{\linewidth}{!}{%
\begin{tabular}{|l|c|c|c|c|c|c|c|}
\hline
\textbf{Method} & \textbf{T1} & \textbf{T2} &
\textbf{T3} & \textbf{T4} & \textbf{T5} &
\textbf{Avg} & \textbf{BWT} \\
\hline
SFT       & 32.3  & 5.8   & 0.0   & 57.9  & 100.0 & 39.20 & -75.90 \\ \hline
RSD-AMC   & 99.22 & 98.45 & 99.88 & 98.64 & 99.94 & 99.23 & -0.81 \\ \hline
EWC       & 88.4  & 95.9  & 99.9  & 87.3  & 100.0 & 94.32 & -7.02 \\ \hline
MSNCIL    & 99.84 & 99.83 & 99.96 & 99.17 & 99.91 & 99.74 & 0.00 \\ \hline
PRISM-AMC & 99.1  & 99.0  & 99.9  & 98.0  & 100.0 & 99.20 & -0.87 \\ \hline
\end{tabular}%
}
\end{table}

\begin{table}[!htbp]
\centering
\footnotesize
\caption{Continual learning performance on the MIGOU dataset under class-disjoint task sequences. Reporting final task accuracies, overall average accuracy (Avg), and backward transfer (BWT).}
\label{tab:migounooverlap}
\renewcommand{\arraystretch}{1.1}
\setlength{\tabcolsep}{4pt}
\begin{tabular}{|c|c|c|c|c|c|c|c|}
\hline
\textbf{Method} & \textbf{T1} & \textbf{T2} & \textbf{T3} & \textbf{T4} & \textbf{T5} & \textbf{Avg} & \textbf{BWT} \\
\hline
\textbf{SFT} & 63.8 & 94.0 & 86.9 & 81.8 & 64.6 & 78.22 & -17.73 \\
\hline
\textbf{RSD-AMC} & 82.9  & 99.52 & 99.9 & 98.2 & 69.2 & 89.94 & -4.55 \\
\hline
\textbf{EWC} & 89.3 & 89.5 & 83.9 & 83.2 & 70.3 & 83.26 & -13.24 \\
\hline
\textbf{MSNCIL} & 98.79  & 99.89 &  99.80  & 89.19 & 50.01 & 87.54 & 0.00 \\
\hline
\textbf{PRISM-AMC} & 89.9 & 99.1 & 99.7 & 99.6 & 59.4 & 89.53 & -2.28 \\
\hline
\end{tabular}
\end{table}


\begin{table}[t]
\centering
\footnotesize
\caption{Time per epoch (seconds) across tasks. Replay exhibits increasing computational cost with task index, whereas PRISM-AMC stabilizes after Task 1.}
\label{tab:time_per_epoch}
\renewcommand{\arraystretch}{1.1}
\setlength{\tabcolsep}{4pt}
\begin{tabular}{|c|c|c|c|c|c|}
\hline
\textbf{Method} & \textbf{T1} & \textbf{T2} & \textbf{T3} & \textbf{T4} & \textbf{T5} \\
\hline
\textbf{RSD-AMC} & 68.03 & 73.83 & 101.43 & 155.15 & 215.92 \\
\hline
\textbf{SFT} & 56.30 & 33.78 & 33.65 & 34.05 & 34.10 \\
\hline
\textbf{MSNCIL} & 60.56 & 42.14 & 42.18 & 42.39 & 42.33 \\ \hline
\textbf{EWC} & 50.70 & 64.06 & 94.81 & 157.03 & 191.12 \\
\hline
\textbf{PRISM-AMC} & 54.18 & 85.96 & 86.05 & 86.41 & 86.39 \\
\hline 
\end{tabular}
\end{table}

\begin{table}[!htbp]
\centering
\caption{Memory growth on MIGOU (float32). PRISM-AMC memory is computed from the retained gradient-constraint subspaces. Replay (RSD-AMC) stores 120 raw I/Q examples per task.}
\label{tab:memory_growth_migou}
\renewcommand{\arraystretch}{1.15}
\setlength{\tabcolsep}{7pt}
\begin{tabular}{|c|c|c|c|c|c|}
\hline
\textbf{\#Tasks} & 1 & 2 & 3 & 4 & 5 \\
\hline
\textbf{PRISM-AMC (MB)} & 2.52 & 4.94 & 6.66 & 8.23 & 9.05 \\
\hline
\textbf{RSD-AMC} & 0.12 & 0.24 & 0.36 & 0.48 & 0.60 \\
\hline 
\textbf{MSNCIL} & 8,52 & 8.52 & 8.52 & 8.52 & 8.52 \\ \hline
\textbf{EWC}& 27.00 & 54.00 & 81.01 & 108.01 & 135.02 \\
\hline
\end{tabular}
\end{table}


\begin{table}[!htbp]
\centering
\footnotesize
\caption{Continual learning performance on the MIGOU dataset under class-disjoint task sequences: Ablation over different values of $\epsilon_{th}$.}
\label{tab:gpm_eps}
\renewcommand{\arraystretch}{1.1}
\setlength{\tabcolsep}{4pt}
\begin{tabular}{|c|c|c|c|c|c|c|c|}
\hline
$\epsilon$ & T1 & T2 & T3 & T4 & T5 & Avg & BWT \\
\hline
0.750  & 41.6 & 77.4 & 75.7 & 74.1 & 74.8 & 68.71 & -32.56 \\
\hline
0.900  & 76.6 & 94.1 & 88.2 & 97.1 & 72.6 & 85.71 & -10.73 \\
\hline
0.925  & 77.2 & 95.3 & 96.1 & 97.4 & 72.5 & 87.67 & -8.25 \\
\hline
0.950  & 75.0 & 92.7 & 99.2 & 97.5 & 69.9 & 86.86 & -8.63 \\
\hline
0.970  & 89.9 & 99.1 & 99.7 & 99.6 & 59.4 & \textbf{89.53} & \textbf{-2.28} \\
\hline
\end{tabular}
\end{table}


\begin{table}[t]
\centering
\footnotesize
\caption{Continual learning performance on the MIGOU dataset under class-disjoint task sequences: Removing bias results in better performance.}
\label{tab:bias}
\renewcommand{\arraystretch}{1.1}
\setlength{\tabcolsep}{4pt}
\begin{tabular}{|c|c|c|c|c|c|c|c|}
\hline
\textbf{Method} & \textbf{T1} & \textbf{T2} & \textbf{T3} & \textbf{T4} & \textbf{T5} & \textbf{Avg} & \textbf{BWT} \\
\hline
With bias term    & 80.9 & \textbf{99.4} & 89.5 & 83.6 & \textbf{70.6} & 84.69 & -11.35 \\
\hline
Without bias term & \textbf{89.9} & 99.1 & \textbf{99.7} & \textbf{99.6} & 59.4 & \textbf{89.53} & \textbf{-2.28} \\
\hline
\end{tabular}
\end{table}

\subsubsection{Single-head 11-way class-incremental setting} We proceed to consider a more challenging yet practical single-head 11-way class-incremental setting on the MIGOU dataset. In contrast to the task-specific-head setting considered above, the single-head class-incremental setting does not provide task identity at test time. Thus, a task-specific classifier cannot be selected during inference. We therefore replace the task-specific output heads with a unified cosine-prototype classifier over all classes observed so far. After each session, prototypes for the newly introduced classes are computed by averaging the normalized feature embeddings of their training examples and are retained for subsequent inference; prototypes of previously observed classes are retained without storing their raw I/Q samples. During inference, a test example is assigned to the class whose prototype has the largest cosine similarity with its embedding. This ensures that all seen classes are compared in a shared output space.

To study the effect of the initial base-session size, we evaluate four class-incremental protocols while keeping the final classification problem fixed at 11 classes. \begin{itemize}
\item \textbf{Setting 1 (9+2):} The base session contains BPSK, QPSK, 8PSK, PAM4, WBFM, QAM16, GFSK, QAM64, and AM-SSB. The second session introduces CPFSK and AM-DSB.

\item \textbf{Setting 2 (7+2+2):} The base session contains BPSK, QPSK, 8PSK, PAM4, WBFM, QAM16, and GFSK. The second session introduces QAM64 and CPFSK. The third session introduces AM-DSB and AM-SSB.

\item \textbf{Setting 3 (5+2+2+2):} The base session contains BPSK, QPSK, 8PSK, PAM4, and WBFM. The second session introduces QAM16 and QAM64. The third session introduces GFSK and CPFSK. The fourth session introduces AM-DSB and AM-SSB.

\item \textbf{Setting 4 (3+2+2+2+2):} The base session contains BPSK, QPSK, and 8PSK. The second session introduces PAM4 and QAM16. The third session introduces QAM64 and GFSK. The fourth session introduces CPFSK and WBFM. The fifth session introduces AM-DSB and AM-SSB.

\end{itemize}

The results, presented in Table~\ref{tab:migou_11way_perclass_all}, show that PRISM-AMC consistently outperforms all competing baselines across the considered class-incremental settings. The strongest baseline varies across settings: EWC performs best in Settings~1 and~4, RSD-AMC in Setting~2, and PASS-Net in Setting~3. PRISM-AMC improves over the strongest competing baseline by $21.15$, $18.17$, $21.07$, and $26.06$ percentage points in Settings~1--4, respectively.

\begin{table*}[t]
\centering
\caption{Per-class final accuracy on the MIGOU dataset under single-head 11-way class-incremental protocols. Each row reports accuracy after completing all sessions for a given method and class-incremental setting.}
\label{tab:migou_11way_perclass_all}
\scriptsize
\begin{tabular}{llcccccccccccc}
\hline
Setting & Method 
& BPSK & QPSK & 8PSK & PAM4 & WBFM & GFSK 
& QAM16 & QAM64 & CPFSK & AM-SSB & AM-DSB 
& Avg  \\
\hline
\textbf{Joint \& One-session} & -- & 99.41 & 99.49 & 92.47 & 99.16 & 63.03 & 56.60 & 99.90 & 99.94 & 87.30 & 72.01 & 52.39 & 83.79\\
\hline
\multirow{5}{*}{\textbf{Setting 1}}
& SFT       &  81.34 & 42.09 & 8.64 & 92.43 & 29.41 & 42.47 & 71.16 & 100.00 & 45.81 & 100.00 & 0.20 & 55.78 \\
& EWC       & 94.30 & 75.80 & 54.60 & 95.90 & 58.10 & 55.70 & 96.30 & 100.00 & 47.70 & 100.00 & 0.30 & 70.79  \\
& RSD-AMC   & 95.5 & 91.7 & 85.1 & 93.4 & 69.9 & 67.5 & 98.9 & 12.9 & 66.7 & 69.8 & 22.8 & 70.38   \\
& PASS-Net  & 79.67 & 83.34 & 94.43 & 70.01 & 32.99 & 0.00 & 78.61 & 80.47 & 67.49 & 15.64 & 0.00 & 54.79 \\
& PRISM-AMC & 99.03 & 92.10 & 95.04 & 98.57 & 77.53 & 69.24 & 99.90 & 100.00 & 80.34 & 100.00 & 99.53 & \textbf{91.94}  \\
\hline

\multirow{5}{*}{\textbf{Setting 2}}
& SFT       &18.16 & 4.09 & 0.51 & 5.77 & 38.70 & 46.70 & 6.73 & 35.70 & 35.19 & 84.24 & 65.14 & 30.99\\
& EWC      & 34.61 & 61.71 & 12.20 & 69.99 & 56.04 & 45.79 & 28.14 & 36.37 & 68.67 & 81.71 & 62.96 & 50.75 \\
& RSD-AMC  & 85.3 & 82.3 & 79.2 & 78.7 & 43.0 & 25.9 & 85.2 & 20.4 & 51.0 & 13.6 & 14.8 & 52.67   \\
& PASS-Net  & 47.87 & 28.74 & 58.71 & 54.37 & 99.93 & 34.60 & 99.00 & 37.91 & 12.07 & 0.01 & 0.13 & 43.03 \\
& PRISM-AMC & 71.46 & 95.83 & 67.07 & 55.53 & 75.17 & 82.69 & 24.99 & 77.81 & 85.70 & 78.59 & 64.40 & \textbf{70.84}\\
\hline
\multirow{5}{*}{\textbf{Setting 3}}
& SFT       &1.84 & 4.63 & 3.04 & 0.96 & 16.51 & 14.31 & 18.53 & 0.24 & 21.53 & 85.23 & 62.26 & 20.83\\
& EWC       & 0.26 & 1.99 & 5.69 & 36.41 & 19.54 & 6.03 & 28.47 & 29.73 & 34.80 & 81.06 & 53.80 & 27.07 \\
& RSD-AMC & 68.1 & 34.7 & 22.1 & 62.7 & 10.7 & 12.1 & 18.2 & 14.3 & 43.9 & 16.3 & 9.5 & 28.42 \\
& PASS-Net &  42.27 & 13.24 & 96.17 & 43.91 & 41.11 & 8.59 & 18.44 & 0.87 & 24.64 & 62.73 & 8.93 & 32.81 \\
& PRISM-AMC & 66.40 & 28.24 & 68.07 & 70.74 & 23.50 & 13.29 & 63.34 & 60.89 & 69.64 & 74.13 & 54.46 & \textbf{53.88} \\
\hline

\multirow{5}{*}{\textbf{Setting 4}}
& SFT       & 23.91 & 24.96 & 12.13 & 15.94 & 6.87 & 6.01 & 11.74 & 23.04 & 8.63 & 84.29 & 60.11 & 25.24 \\
& EWC       & 2.54 & 16.47 & 3.23 & 40.34 & 28.59 & 37.37 & 24.01 & 0.30 & 10.09 & 83.16 & 54.43 & 27.32 \\
& RSD-AMC   & 41.2 & 17.7 & 15.3 & 36.3 & 7.9 & 19.8 & 50.7 & 28.0 & 16.5 & 10.9 & 15.1 & 23.58 \\
& PASS-Net & 99.60 & 2.64 & 0.00 & 23.57 & 0.00 & 0.00 & 2.24 & 100.00 & 3.14 & 0.49 & 0.01 & 21.06 \\
& PRISM-AMC & 59.00 & 34.90 & 60.71 & 42.67 & 50.40 & 55.29 & 69.10 & 65.47 & 22.07 & 76.21 & 51.39 & \textbf{53.38} \\
\hline

\end{tabular}
\end{table*}

\subsection{Results on DeepSig 2018.01A Dataset}

We conduct a second set of experiments on the DeepSig 2018.01A dataset~\cite{deepsig2018}, which contains 24 digital and analog modulation classes with SNR values ranging from $-20$\,dB to $30$\,dB in 2\,dB increments. Each sample consists of $1024$ complex I/Q symbols, and for every modulation–SNR pair the dataset provides $4096$ examples.

We consider a class-incremental scenario with $T=5$ tasks and two settings, with $7$ and $15$ randomly selected modulation classes per task, respectively. For each task, the training set contains $85{,}000$ samples per class aggregated across the SNR range, while the validation and test sets contain $5{,}000$ and $15{,}000$ samples per class, respectively.

\begin{figure}[!htbp]
    \centering    \includegraphics[width=\columnwidth]{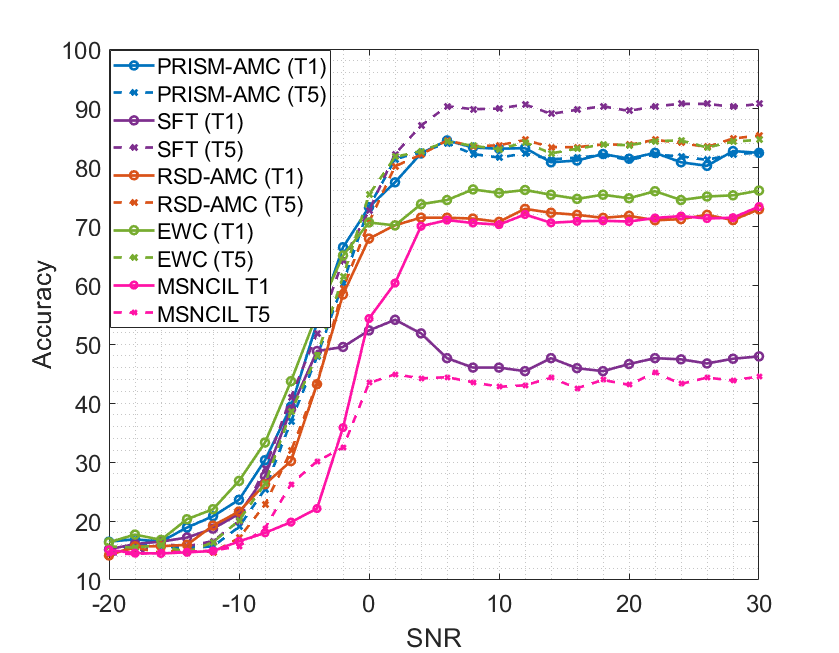}
    \vspace{-2mm}
    \caption{SNR-dependent accuracy on the DeepSig dataset (7 classes per task) after training up to Task~5. For clarity, only the curves for Tasks~1 and~5 are shown.}
    \hfill
    \label{fig:Deepsigres7}
\end{figure}

\begin{figure}[!htbp]
    \centering    \includegraphics[width=\columnwidth]{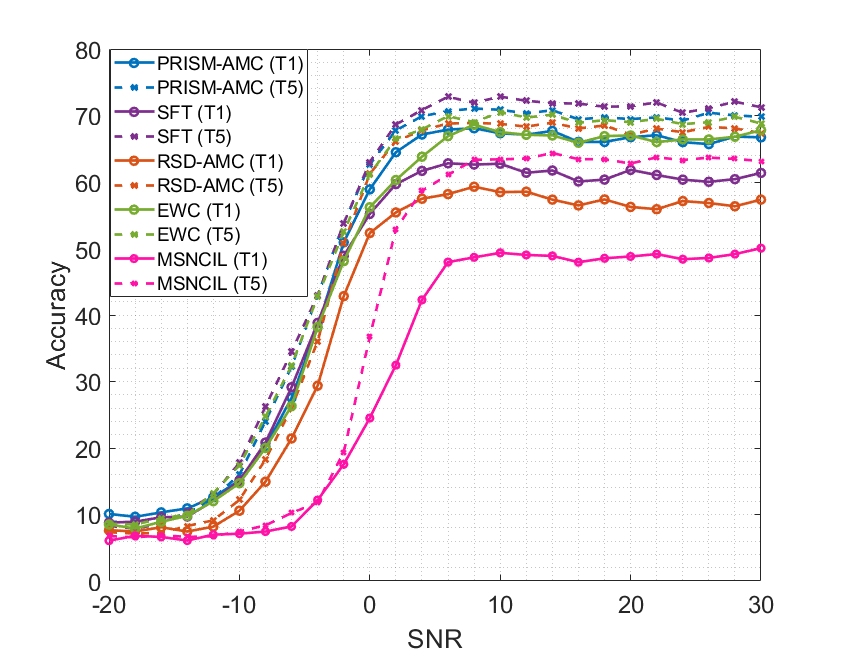}
    \vspace{-2mm}
    \caption{SNR-dependent accuracy on the DeepSig dataset (15 classes per task) after training up to Task~5. For clarity, only the curves for Tasks~1 and~5 are shown.}
    \hfill
    \label{fig:Deepsigres}
\end{figure}

Table~\ref{tab:deepsig_final_comparison7} and Table~\ref{tab:deepsig_final_comparison15} report the final SNR-averaged task accuracies after training on all five tasks. The SNR-dependent performance for Tasks~1 and~5 in the 7-class and 15-class settings is shown in Fig.~\ref{fig:Deepsigres7} and Fig.~\ref{fig:Deepsigres}, respectively. In the 7-class setting, all competing baselines experience substantial catastrophic forgetting, with PRISM-AMC exceeding the best baseline by $8.45$ percentage points in average accuracy and by $9.35$ percentage points in BWT. In the 15-class setting, EWC and PRISM-AMC both retain strong performance on Task~1, particularly in the high-SNR regime. On Task~5, however, EWC slightly underperforms both PRISM-AMC and SFT in the high-SNR regime.


\begin{table}[!htbp]
\centering
\footnotesize
\caption{Continual learning performance on the DeepSig 2018.01A dataset (7 classes per task). Final task accuracies are averaged across SNR values; overall average accuracy (Avg) and backward transfer (BWT) are also reported.}
\label{tab:deepsig_final_comparison7}
\renewcommand{\arraystretch}{1.1}
\setlength{\tabcolsep}{4pt}
\begin{tabular}{|c|c|c|c|c|c|c|c|}
\hline
\textbf{Method} & \textbf{T1} & \textbf{T2} & \textbf{T3} & \textbf{T4} & \textbf{T5} & \textbf{Avg} & \textbf{BWT} \\
\hline
\textbf{SFT} & 38.45 & 35.82 & 25.63 & 26.19 & 66.74 & 38.52 & -28.47 \\
\hline
\textbf{RSD-AMC} & 47.54 & 53.98 & 41.74 & 39.37 & 59.45 & 48.36 & -12.75 \\
\hline
\textbf{EWC} & 53.03 & 48.45 & 38.64 & 40.85 & 62.42 & 48.65 & -14.15 \\
\hline
\textbf{PRISM-AMC} & \textbf{61.39} & \textbf{59.56} & \textbf{53.49} & \textbf{48.95} & 62.13 & \textbf{57.10} & \textbf{-3.40} \\
\hline
\end{tabular}
\end{table}


\begin{table}[!htbp]
\centering
\footnotesize
\caption{Continual learning performance on the DeepSig 2018.01A dataset (15 classes per task). Final task accuracies are averaged across SNR values; overall average accuracy (Avg) and backward transfer (BWT) are also reported.}
\label{tab:deepsig_final_comparison15}
\renewcommand{\arraystretch}{1.1}
\setlength{\tabcolsep}{4pt}
\begin{tabular}{|c|c|c|c|c|c|c|c|}
\hline
\textbf{Method} & \textbf{T1} & \textbf{T2} & \textbf{T3} & \textbf{T4} & \textbf{T5} & \textbf{Avg} & \textbf{BWT} \\
\hline
\textbf{SFT} & 46.7 & 46.5 & 50.0 & 47.8 & 56.0 & 49.40 & -2.68 \\
\hline
\textbf{RSD-AMC} & 41.0 & 41.0 & 43.2 & 36.9 & 50.4 & 42.50 & -6.68 \\
\hline
\textbf{EWC} & 50.3 & 48.0 & 50.1 & 43.8 & 53.0 & 49.02 & -0.51 \\
\hline
\textbf{PRISM-AMC} & \textbf{51.1} & \textbf{49.1} & \textbf{50.6} & 45.6 & 54.4 & \textbf{50.16} & \textbf{-0.30} \\
\hline
\end{tabular}
\end{table}

We then proceed to consider four single-head 24-way class-incremental learning settings that differ in the number of modulation classes included in the initial session and the number of subsequent incremental sessions:

\begin{itemize}
\item \textbf{Setting 1:} The initial session contains 22 modulation classes. The remaining two classes, \text{OQPSK} and \text{16APSK}, are introduced together in one incremental session.

\item \textbf{Setting 2:} The initial session contains 20 modulation classes. Two additional incremental sessions are then introduced, containing \{\text{AM-DSB-WC}, \text{OQPSK}\} and \{\text{32PSK}, \text{16APSK}\}, respectively.

\item \textbf{Setting 3:} The initial session contains 18 modulation classes. The remaining classes are introduced over three incremental sessions containing \{\text{16PSK}, \text{32PSK}\}, \{\text{16APSK}, \text{64APSK}\}, and \{\text{AM-DSB-WC}, \text{OQPSK}\}, respectively.

\item \textbf{Setting 4:} The initial session contains 16 modulation classes. Four subsequent incremental sessions introduce two classes each: \{\text{16PSK}, \text{32PSK}\}, \{\text{16APSK}, \text{64APSK}\}, \{\text{256QAM}, \text{AM-DSB-WC}\}, and \{\text{AM-DSB-SC}, \text{OQPSK}\}, respectively.

\end{itemize}

\begin{figure}[!t]
    \centering
    \includegraphics[width=\columnwidth]
    {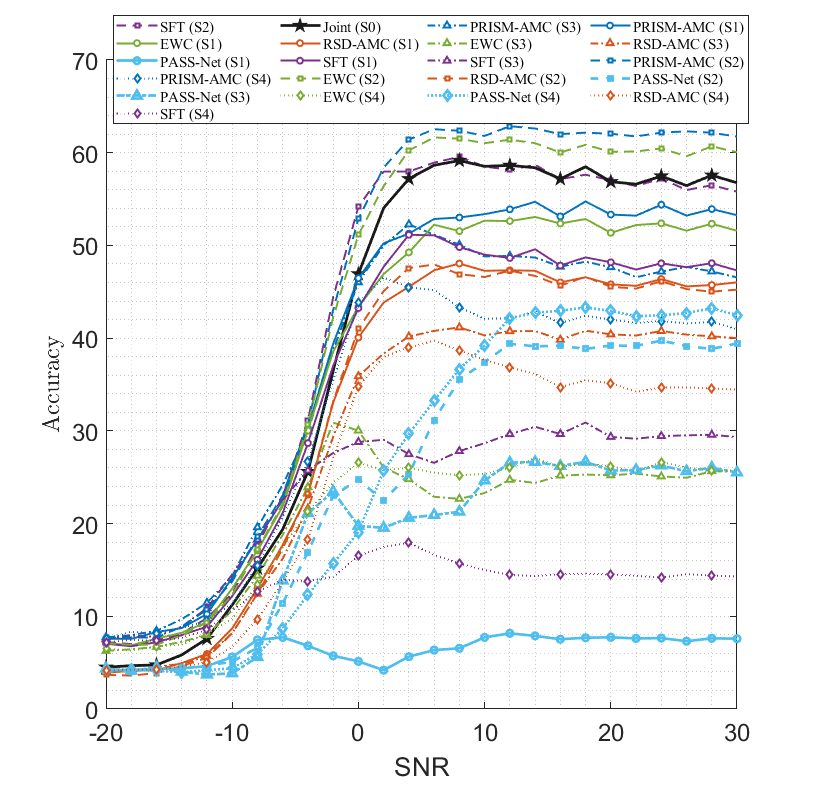}
    \caption{SNR-dependent classification accuracy on the DeepSig 2018.01A
    dataset under different single-head 24-way class-incremental settings. Setting~0 denotes
    joint training using all modulation classes, while Settings~1--4 represent
    class-incremental configurations with an increasing number of incremental
    sessions. PRISM-AMC consistently outperforms the
    competing baselines across Settings~1--4.}
    \label{fig:deepsig_class_incremental_settings}
\end{figure}

For comparison, we also consider a non-incremental setting in which all 24 modulation classes are available during a single training session. As shown in Fig.~\ref{fig:deepsig_class_incremental_settings}, PRISM-AMC consistently outperforms the competing methods across the class-incremental settings, further demonstrating its effectiveness in mitigating catastrophic forgetting.

\subsection{Results for Domain-Incremental Setting} To investigate the domain-incremental setting, we generate data using \eqref{sysmodeq}. The dataset consists of four modulation classes (BPSK, QPSK, 8-PSK, and 16-QAM), with $35{,}000$ examples per class. All tasks consist of the same modulation classes with equal class proportions, while the signals corresponding to each class differ across tasks according to the underlying channel statistics. All tasks use a common classifier head. We model the amplitude and phase offsets as $A \sim \mathrm{Rayleigh}(A_u)$ and $\Delta\theta \sim \mathcal{U}(0,\theta_u)$, respectively, where $A_u$ is the Rayleigh scale parameter.

\begin{figure}[htbp]
    \centering
    \subfloat[Tasks 1--3]{
        \includegraphics[width=\columnwidth]{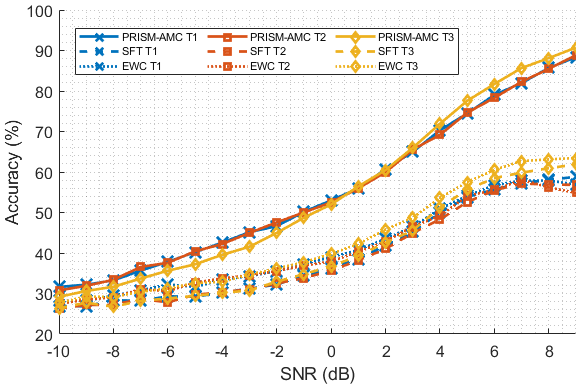}
        \label{fig:DomIncCFO_T1to3}
    }

    \vspace{1mm}

    \subfloat[Tasks 4--7]{
        \includegraphics[width=\columnwidth]{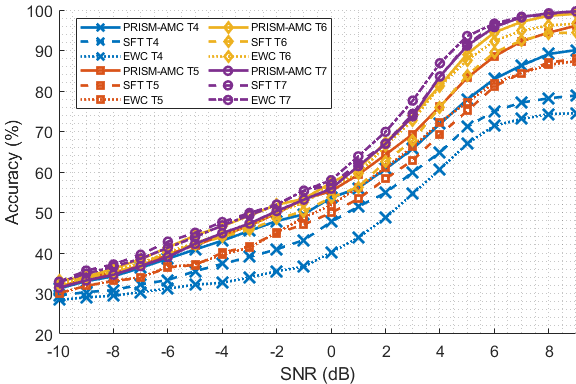}
        \label{fig:DomIncCFO_T4to7}
    }

    \vspace{-2mm}
        \caption{SNR-dependent accuracy for the domain-incremental setting with CFO varying across tasks.}
    \label{fig:DomIncCFO}
\end{figure}

For the first set of experiments, we set $A_u=0.35$ and $\theta_u=\pi/16$, and vary $\Delta f$ across tasks, with $\Delta f=[0,0.01,0.1,0.5,1.0,10,20]$ for Tasks~1 through~7, respectively. The results presented in Fig.~\ref{fig:DomIncCFO} and Tables~\ref{tab:prism_cfo_results} and \ref{tab:domain_incremental_cfo} show that PRISM-AMC substantially reduces catastrophic forgetting under changes in carrier frequency offset. At SNR $=10$\,dB, PRISM-AMC achieves a final average accuracy of $94.54\%$, compared with $77.31\%$ for SFT and $76.09\%$ for EWC, while improving BWT from $-26.26$ and $-27.70$ to $-6.11$, respectively.

\begin{table}[t]
\centering
\caption{Task-wise accuracy (\%) matrix for PRISM-AMC under the
domain-incremental setting with carrier frequency offset varying across
tasks, evaluated at $\mathrm{SNR}=10$\,dB. Entry $(i,j)$ denotes the
accuracy on task $j$ after training up to task $i$. Final average accuracy:
$94.54\%$; BWT: $-6.11$.}
\label{tab:prism_cfo_results}
\renewcommand{\arraystretch}{1.2}
\setlength{\tabcolsep}{4pt}
\begin{tabular}{|c|c|c|c|c|c|c|c|}
\hline
 & \multicolumn{7}{c|}{\textbf{Evaluation Task}} \\
\hline
\textbf{After Task} & \textbf{T1} & \textbf{T2} & \textbf{T3} & \textbf{T4} & \textbf{T5} & \textbf{T6} & \textbf{T7} \\
\hline
T1 & 99.7 & -- & -- & -- & -- & -- & --  \\\hline
T2 & 99.7 & 99.7 & -- & -- & -- & -- & -- \\\hline
T3 & 99.6 & 99.7 & 99.7 & -- & -- & -- & -- \\\hline
T4 & 98.3 & 98.2 & 98.2 & 99.9 & -- & -- & -- \\\hline
T5 & 96.3 & 96.4 & 95.6 & 98.0 & 99.9 & -- & -- \\\hline
T6 & 94.3 & 93.9 & 93.4 & 94.8 & 98.5 & 99.8 & -- \\\hline
T7 & 91.2 & 90.7 & 92.1 & 91.5 & 97.2 & 99.2 & 99.8 \\
\hline
\end{tabular}

\vspace{2mm}

\end{table}

\begin{table}[!t]
\centering
\caption{Continual learning performance under the
domain-incremental setting (Carrier Frequency Offset),
evaluated at SNR = 10 dB. Reporting final task accuracies
after training on all seven tasks, final average accuracy
(Avg), and backward transfer (BWT).}
\label{tab:domain_incremental_cfo}
\resizebox{\linewidth}{!}{%
\begin{tabular}{|l|c|c|c|c|c|c|c|c|c|}
\hline
\textbf{Method} & \textbf{T1} & \textbf{T2} &
\textbf{T3} & \textbf{T4} & \textbf{T5} &
\textbf{T6} & \textbf{T7} & \textbf{Avg} &
\textbf{BWT} \\
\hline
SFT &
59.5 & 56.3 & 61.5 & 79.6 & 89.8 & 94.6 & 99.9 &
77.31 & -26.26 \\ \hline
EWC &
56.5 & 53.5 & 62.8 & 74.8 & 88.3 & 96.8 & 99.9 &
76.09 & -27.70 \\ \hline
PRISM-AMC &
91.2 & 90.7 & 92.1 & 91.5 & 97.2 & 99.2 & 99.8 &
\textbf{94.54} & \textbf{-6.11} \\ \hline

\end{tabular}%
}
\end{table}

\begin{figure}[!htbp]
    \centering
    \subfloat[Tasks 1--2]{
        \includegraphics[width=\columnwidth]{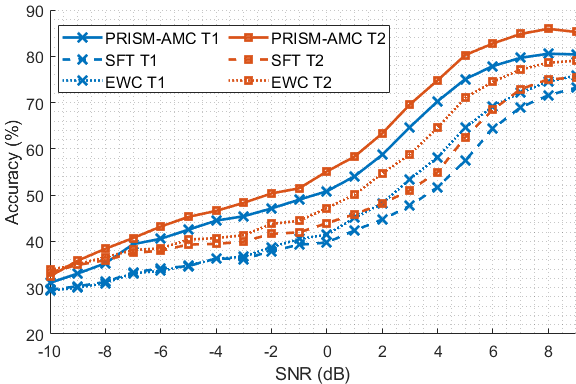}
        \label{fig:DomInctheta_T1to2}
    }

    \vspace{1mm}

    \subfloat[Tasks 3--5]{
        \includegraphics[width=\columnwidth]{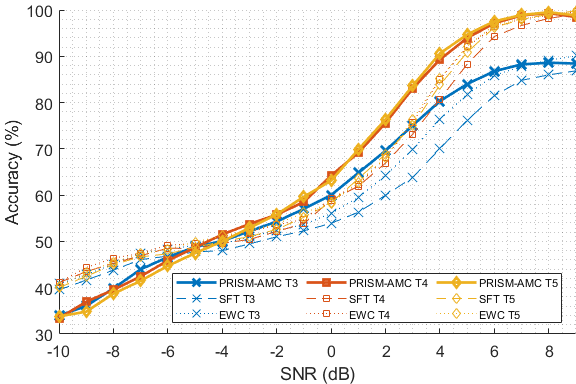 }
        \label{fig:DomInctheta_T3to5}
    }

    \vspace{-2mm}
    \caption{SNR-dependent accuracy for the domain incremental setting ($\theta_u$ varied across tasks) Top: Tasks 1--2. Bottom: Tasks 3--5.}
    \label{fig:DomInctheta}
\end{figure}

To explore variation in phase offset, we fix $\Delta f=0.1$ and
$A_u=0.35$, and vary $\theta_u$ across five tasks as
$[\pi, \frac{\pi}{2}, \frac{\pi}{4}, \frac{\pi}{8}, \frac{\pi}{16}]$.
The results in Fig.~\ref{fig:DomInctheta} and
Tables~\ref{tab:prism_theta_var} and \ref{tab:domain_incremental_theta}
show that PRISM-AMC achieves higher final average accuracy and less
forgetting than EWC and sequential fine-tuning.

\begin{figure}[!htbp]
    \centering
    \subfloat[Tasks 1--2]{
        \includegraphics[width=\columnwidth]{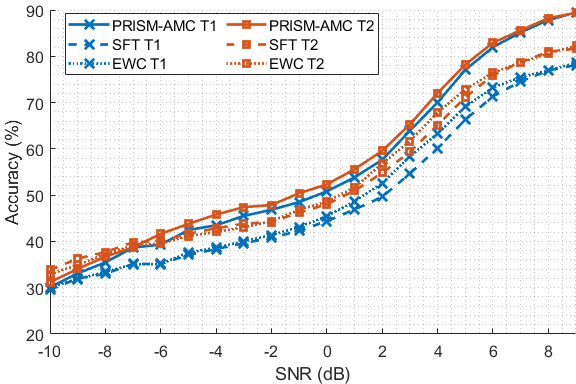}
        \label{fig:DomIncAtheta_T1to2}
    }

    \vspace{1mm}

    \subfloat[Tasks 3--5]{
        \includegraphics[width=\columnwidth]{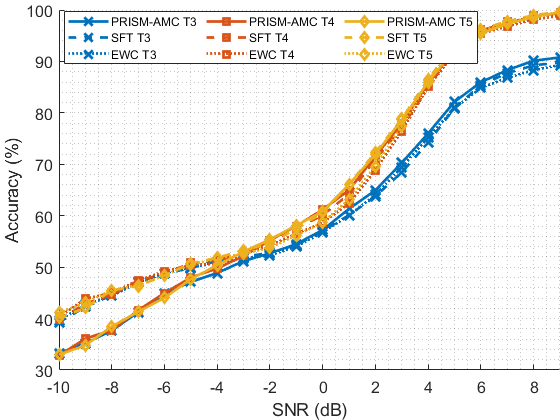 }
        \label{fig:DomIncAtheta_T3to5}
    }

    \vspace{-2mm}
    \caption{SNR-dependent accuracy for the domain incremental setting ($A_u$ and $\theta_u$ jointly varied across tasks) Top: Tasks 1--2. Bottom: Tasks 3--5.}
    \label{fig:DomIncATheta}
\end{figure}

Finally, with $\Delta f=0.1$ fixed, we jointly vary $A_u$ and $\theta_u$ across five tasks using the pairs
\[
\left\{
(1,\pi),
\left(0.75,\frac{\pi}{2}\right),
\left(0.5,\frac{\pi}{4}\right),
\left(0.35,\frac{\pi}{8}\right),
\left(0.25,\frac{\pi}{16}\right)
\right\}.
\]
The corresponding results, presented in Fig.~\ref{fig:DomIncATheta} and Tables~\ref{tab:prism_A_theta_var} and \ref{tab:domain_incremental_joint}, again show that PRISM-AMC achieves higher final average accuracy and less forgetting than EWC and sequential fine-tuning.

\begin{table}[!htbp]
\centering
\caption{Task-wise accuracy (\%) matrix for PRISM-AMC under the domain-incremental setting with $\theta_u$ varied across tasks, evaluated at $\mathrm{SNR}=10$\,dB. Entry $(i,j)$ denotes the accuracy on task $j$ after training up to task $i$. Final average accuracy: $89.48\%$;
BWT: $-3.91$.}
\label{tab:prism_theta_var}
\resizebox{0.48\textwidth}{!}{
\begin{tabular}{|c|c|c|c|c|c|}
\hline
 & \multicolumn{5}{c|}{\textbf{Evaluation Task}} \\
\hline
\textbf{After Task} & \textbf{T1} & \textbf{T2} & \textbf{T3} & \textbf{T4} & \textbf{T5} \\
\hline
T1 & 74.5 & -    & -    & -    & -    \\\hline
T2 & 91.6 & 96.4 & -    & -    & -    \\\hline
T3 & 88.8 & 93.5 & 96.4 & -    & -    \\\hline
T4 & 82.7 & 88.2 & 92.4 & 97.9 & -    \\\hline
T5 & 79.6 & 84.6 & 87.9 & 97.5 & 97.8 \\
\hline
\end{tabular}
}
\end{table}

\begin{table}[!htbp]
\centering
\caption{Continual learning performance under the domain-incremental setting with $\theta_u$ varied across tasks, evaluated at $\mathrm{SNR}=10$\,dB. Final task accuracies after training on all five tasks, final average accuracy (Avg), and backward transfer (BWT) are reported.}
\label{tab:domain_incremental_theta}
\resizebox{\linewidth}{!}{%
\begin{tabular}{|l|c|c|c|c|c|c|c|}
\hline
\textbf{Method} & \textbf{T1} & \textbf{T2} &
\textbf{T3} & \textbf{T4} & \textbf{T5} &
\textbf{Avg} & \textbf{BWT} \\
\hline
SFT &
72.3 & 75.4 & 86.3 & 98.7 & 99.3 &
86.40 & -9.85 \\ \hline
EWC &
75.6 & 79.0 & 90.3 & 99.4 & 99.5 &
88.76 & -6.98 \\ \hline
PRISM-AMC &
79.6 & 84.6 & 87.9 & 97.5 & 97.8 &
\textbf{89.48} & \textbf{-3.91} \\ \hline

\end{tabular}%
}
\end{table}

\begin{table}[!htbp]
\centering
\caption{Task-wise accuracy (\%) matrix for PRISM-AMC under the domain-incremental setting with $A_u$ and $\theta_u$ jointly varied across tasks, evaluated at $\mathrm{SNR}=10$\,dB. Entry $(i,j)$ denotes the accuracy on task $j$ after training up to task $i$. Final average accuracy: $93.62\%$; BWT: $-5.69$.}
\label{tab:prism_A_theta_var}
\resizebox{0.48\textwidth}{!}{
\begin{tabular}{|c|c|c|c|c|c|}
\hline
 & \multicolumn{5}{c|}{\textbf{Evaluation Task}} \\
\hline
\textbf{After Task} & \textbf{T1} & \textbf{T2} & \textbf{T3} & \textbf{T4} & \textbf{T5} \\
\hline
T1 & 94.1 & -    & -    & -    & -    \\\hline
T2 & 98.7 & 99.3 & -    & -    & -    \\\hline
T3 & 97.2 & 98.1 & 99.0 & -    & -    \\\hline
T4 & 94.5 & 94.9 & 97.6 & 99.2 & -    \\\hline
T5 & 89.3 & 89.5 & 90.8 & 99.3 & 99.2 \\
\hline
\end{tabular}
}
\end{table}

\begin{table}[!htbp]
\centering
\caption{Continual learning performance under the domain-incremental setting with $A_u$ and $\theta_u$ jointly varied across tasks, evaluated at $\mathrm{SNR}=10$\,dB. Final task accuracies after training on all five tasks, final average accuracy (Avg), and backward transfer (BWT) are reported.}
\label{tab:domain_incremental_joint}
\resizebox{\linewidth}{!}{%
\begin{tabular}{|l|c|c|c|c|c|c|c|}
\hline
\textbf{Method} & \textbf{T1} & \textbf{T2} &
\textbf{T3} & \textbf{T4} & \textbf{T5} &
\textbf{Avg} & \textbf{BWT} \\
\hline
SFT &
78.9 & 81.9 & 89.8 & 99.3 & 99.5 &
89.88 & -10.49 \\ \hline
EWC &
79.7 & 82.3 & 89.5 & 99.2 & 99.6 &
90.08 & -10.26 \\ \hline
PRISM-AMC &
89.3 & 89.5 & 90.8 & 99.3 & 99.2 &
\textbf{93.62} & \textbf{-5.69} \\
\hline
\end{tabular}%
}
\end{table}

\section{Conclusion}

We presented PRISM-AMC, a projection-based continual learning framework for automatic modulation classification under raw-sample retention constraints. By retaining task-relevant activation subspaces rather than raw I/Q samples, PRISM-AMC mitigates catastrophic forgetting without replay. We further analyzed the projected stochastic-gradient updates under an SNR-dependent gradient-noise model. Experiments in both class-incremental and domain-incremental settings demonstrate reduced forgetting and competitive or superior accuracy relative to sequential, regularization-based, and replay-based baselines.


\label{conc}

\bibliographystyle{IEEEbib}
\bibliography{refs}


\newpage
\appendices

\section{Detection of a New Modulation Class}
\label{sec:new_class_detection}

\begin{table*}[!htbp]
\centering
\caption{Epistemic uncertainty obtained using MC dropout in leave-one-class-out experiments. For each row, the model is trained with the indicated modulation class excluded, and each entry reports the mean epistemic uncertainty for test samples from the corresponding column class. The diagonal entries, shown in bold, therefore correspond to the held-out class. The final column reports the AUROC for discriminating held-out-class samples from samples belonging to the known classes using epistemic uncertainty.}
\label{tab:epistemic_uncertainty}
\setlength{\tabcolsep}{3.2pt}
\renewcommand{\arraystretch}{1.15}
\resizebox{\textwidth}{!}{
\begin{tabular}{lcccccccccccc}
\toprule
\textbf{Held-out class}
& \textbf{BPSK}
& \textbf{QPSK}
& \textbf{8PSK}
& \textbf{PAM4}
& \textbf{WBFM}
& \textbf{GFSK}
& \textbf{QAM16}
& \textbf{QAM64}
& \textbf{CPFSK}
& \textbf{AM-SSB}
& \textbf{AM-DSB}
& \textbf{AUROC} \\
\midrule

BPSK
& \textbf{0.380} & 0.026 & 0.054 & 0.008 & 0.092 & 0.092
& 0.010 & 0.003 & 0.034 & 0.032 & 0.037 & 0.971 \\

QPSK
& 0.010 & \textbf{0.044} & 0.039 & 0.013 & 0.090 & 0.091
& 0.007 & 0.003 & 0.033 & 0.030 & 0.033 & 0.631 \\

8PSK
& 0.014 & 0.025 & \textbf{0.222} & 0.011 & 0.087 & 0.079
& 0.009 & 0.004 & 0.033 & 0.034 & 0.031 & 0.935 \\

PAM4
& 0.009 & 0.027 & 0.059 & \textbf{0.481} & 0.096 & 0.089
& 0.016 & 0.003 & 0.039 & 0.040 & 0.043 & 0.967 \\

WBFM
& 0.010 & 0.025 & 0.041 & 0.012 & \textbf{0.038} & 0.030
& 0.009 & 0.005 & 0.026 & 0.031 & 0.031 & 0.491 \\

GFSK
& 0.011 & 0.031 & 0.064 & 0.012 & 0.032 & \textbf{0.044}
& 0.008 & 0.004 & 0.035 & 0.035 & 0.041 & 0.508 \\

QAM16
& 0.012 & 0.027 & 0.050 & 0.012 & 0.094 & 0.090
& \textbf{0.030} & 0.004 & 0.023 & 0.028 & 0.027 & 0.484 \\

QAM64
& 0.012 & 0.025 & 0.057 & 0.015 & 0.092 & 0.086
& 0.013 & \textbf{0.197} & 0.036 & 0.035 & 0.041 & 0.916 \\

CPFSK
& 0.012 & 0.023 & 0.058 & 0.009 & 0.084 & 0.083
& 0.003 & 0.002 & \textbf{0.142} & 0.006 & 0.008 & 0.685 \\

AM-SSB
& 0.011 & 0.025 & 0.052 & 0.014 & 0.093 & 0.091
& 0.007 & 0.004 & 0.036 & \textbf{0.028} & 0.020 & 0.540 \\

AM-DSB
& 0.011 & 0.025 & 0.051 & 0.013 & 0.097 & 0.089
& 0.009 & 0.005 & 0.035 & 0.024 & \textbf{0.035} & 0.532 \\

\midrule
\multicolumn{12}{r}{\textbf{Average AUROC}}
& \textbf{0.696} \\
\bottomrule
\end{tabular}
}
\end{table*}

In this section, we investigate the detection of wireless signals belonging to modulation classes that were not observed during training. To this end, we employ Monte Carlo (MC) dropout \cite{gal2016dropout} to estimate the model's epistemic uncertainty. In \cite{gal2016dropout}, it is shown that a neural network trained with dropout can be interpreted as an approximate Bayesian neural network. Consequently, stochastic samples from an approximate posterior distribution over the model parameters can be obtained by retaining dropout during inference and performing multiple forward passes.

As discussed in \cite{NIPS2017_2650d608}, predictive uncertainty can broadly be categorized into the following two types:
\begin{itemize}
    \item \textbf{Aleatoric uncertainty:} This uncertainty arises from inherent ambiguity or noise in the observations, such as channel noise, interference, hardware impairments, or overlapping signal characteristics. Under a fixed data-generation and measurement process, aleatoric uncertainty cannot generally be reduced merely by collecting additional training samples.

    \item \textbf{Epistemic uncertainty:} This uncertainty arises from insufficient knowledge of the model parameters due to limited or unrepresentative training data. It is typically elevated for inputs lying outside the training distribution, including signals belonging to previously unseen modulation classes or domains. Unlike aleatoric uncertainty, epistemic uncertainty can generally be reduced by training the model on representative data from the newly encountered classes or domains.
\end{itemize}

Let $\mathbf{r}$ denote an input signal and let $M$ denote the number of
stochastic forward passes. During the $m$-th forward pass, a different
dropout mask is sampled, producing the class-probability vector
\begin{equation}
    \mathbf{p}_m
    =
    p(y\mid\mathbf{r},\boldsymbol{W}_m)
    =
    \operatorname{softmax}
    \left(
        f_{\boldsymbol{W}_m}(\mathbf{r})
    \right),
\end{equation}
where $\boldsymbol{W}_m$ denotes the effective model parameters induced
by the $m$-th dropout realization. The Monte Carlo estimate of the
posterior predictive distribution is
\begin{equation}
    \bar{\mathbf{p}}
    =
    \frac{1}{M}\sum_{m=1}^{M}\mathbf{p}_m.
\end{equation}

To quantify epistemic uncertainty, we compute the mutual information between the predicted class label and the model parameters
\cite{gal2017deep}:
\begin{equation}
\begin{split}
    \mathcal{U}_{\mathrm{epi}}(\mathbf{r})
    &=
    H(\bar{\mathbf{p}})
    -
    \frac{1}{M}\sum_{m=1}^{M}H(\mathbf{p}_m)\\
    &=
    -\sum_{c=1}^{C}\bar{p}_{c}\log\bar{p}_{c}
    +
    \frac{1}{M}\sum_{m=1}^{M}
    \sum_{c=1}^{C}p_{m,c}\log p_{m,c}.
\end{split}
\label{eq:mc_dropout_epistemic}
\end{equation}
where $C$ denotes the total number of modulation classes known to the model. A received signal may be flagged as potentially belonging to an unseen modulation class when
\begin{equation}
    \mathcal{U}_{\mathrm{epi}}(\mathbf{r})>\tau_{\mathrm{det}},
\end{equation}
for a suitably chosen detection threshold $\tau_{\mathrm{det}}$.

We set $M=50$ and evaluate this strategy on the MIGOU dataset using leave-one-class-out experiments. The results in Table~\ref{tab:epistemic_uncertainty} report the mean epistemic uncertainty obtained when each modulation class is excluded during training. The MC-dropout-based detector achieves an average AUROC of $0.696$ across the eleven experiments, indicating moderate overall ability to distinguish previously unseen modulation classes from known classes. Detection performance is strongly class-dependent. When BPSK, PAM4, 8PSK, and QAM64 are held out, the corresponding AUROC values are $0.971$, $0.967$, $0.935$, and $0.916$, respectively. In contrast, performance is close to random chance when WBFM, GFSK, QAM16, AM-SSB, and AM-DSB are held out, with AUROC values ranging from $0.484$ to $0.540$. In these cases, the mean uncertainty of the held-out class is not clearly separated from that of the known classes. This class-dependent behavior motivates combining epistemic uncertainty with an additional feature-space novelty measure, such as the distance or cosine similarity between the received-signal representation and the known-class prototypes.

\end{document}